\documentclass[11pt,a4paper]{article}
\usepackage[hyperref]{AACL-IJCNLP2020}
\usepackage{times}
\usepackage{latexsym}

\usepackage{amsmath}

\usepackage[algo2e, ruled, lined, linesnumbered]{algorithm2e}
\usepackage{multicol,lipsum}
\usepackage{cuted, nccmath}
\usepackage{multirow}
\usepackage{tabularx}
\usepackage{booktabs} 
\usepackage{soul}
\usepackage{balance}
\usepackage{multirow}
\usepackage{graphicx}
\usepackage{amssymb}

\usepackage{subcaption}

\usepackage{epstopdf}
\usepackage{xstring}

\usepackage{epstopdf}
\usepackage[latin1]{inputenc}
\usepackage{xstring}
\usepackage{microtype}

\aclfinalcopy 

\title{Massively Multilingual Document Alignment with Cross-lingual Sentence-Mover's Distance}

\author{Ahmed El-Kishky\\
  Facebook AI \\
  \texttt{ahelk@fb.com} \\\And
  Francisco Guzm\'an \\
  Facebook AI \\
  \texttt{fguzman@fb.com} \\}

\date{}

\begin{document}
\maketitle
\begin{abstract}
Document alignment aims to identify pairs of documents in two distinct languages that are of comparable content or translations of each other. Such aligned data can be used for a variety of NLP tasks from training cross-lingual representations to mining parallel data for machine translation. In this paper we develop an unsupervised scoring function that leverages cross-lingual sentence embeddings to compute the semantic distance between documents in different languages. These semantic distances are then used to guide a document alignment algorithm to properly pair cross-lingual web documents across a variety of low, mid, and high-resource language pairs. Recognizing that our proposed scoring function and other state of the art methods are computationally intractable for long web documents, we utilize a more tractable greedy algorithm that performs comparably. We experimentally demonstrate that our distance metric performs better alignment than current baselines outperforming them by 7\% on high-resource language pairs, 15\% on mid-resource language pairs, and 22\% on low-resource language pairs.
\end{abstract}

\section{Introduction}
While the Web provides a large amount of monolingual text, cross-lingual parallel data is more difficult to obtain. Despite its scarcity, parallel cross-lingual data plays a crucial role in a variety of tasks in natural language processing such as machine translation. Previous works have shown that training on sentences extracted from parallel or comparable documents mined from the Web can improve machine translation models~\cite{munteanu2005improving} or learning
word-level translation lexicons~\cite{fung1998ir,rapp1999automatic}. Other tasks that leverage these parallel texts include cross-lingual information retrieval, document classification, and multilingual representations such as XLM~\cite{lample2019cross}.
\begin{figure}[t]
    \centering
    \includegraphics[width=2.5in]{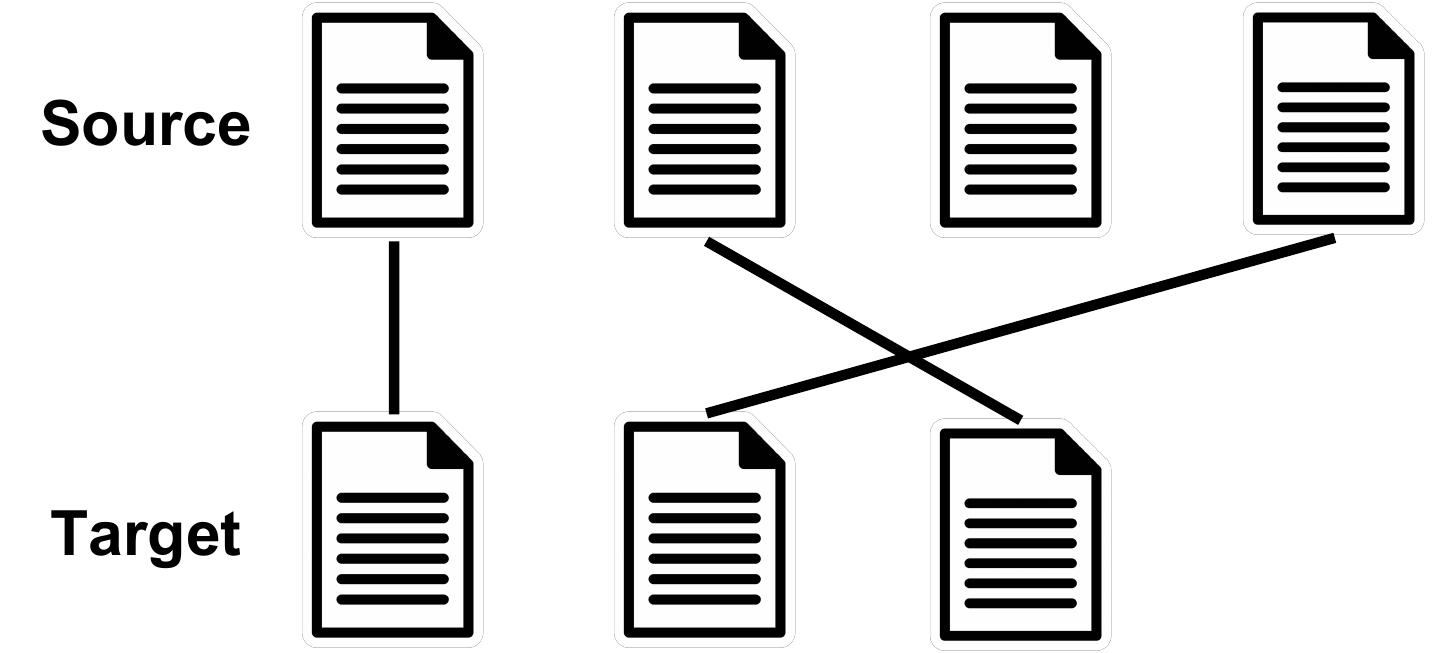}
    \caption{Documents in a source and target langauge in the same web-domain. Solid lines indicate cross-lingual document pairs.}
    \label{fig:aligned_docs}
\end{figure}
Document alignment is a method for obtaining cross-lingual parallel data that seeks to pair documents in different languages such that pairs are translations or near translations of each other. As seen in Figure~\ref{fig:aligned_docs}, this involves a one-to-one pairing of documents in a source language with documents in a target language. 

To automate and scale the process of identifying these documents pairs, we introduce an approach to accurately mine comparable web documents across a variety of low, mid, and high-resource language directions. Previous approaches have been applied to homogeneous corpora, however mining the Web involves analyzing a variety of heterogeneous data sources~\cite{koehn2002europarl}. Other approaches rely on corpus-specific features such as metadata and publication date which can be inconsistent and unreliable~\cite{munteanu2005improving,abdui2009use}. Related methods utilize document structure when calculating document similarity~\cite{resnik2003web,chen2000parallel}. However, when mining large, unstructured collections of web documents these features are often missing or unreliable. As such, we introduce an approach that aligns documents based solely on semantic distances between their textual content.

For our approach, we first decompose documents into sentences, and encode each sentence into a cross-lingual semantic space yielding a bag-of-sentences representation. Utilizing the dense, cross-lingual representation of sentences, we then compute document distances using a variant of earth mover's distance where probability mass is moved from the source document to the target document. We then leverage these document distances as a guiding metric for identifying cross-lingual document pairs and demonstrate experimentally that our proposed method outperforms state-of-the-art baselines that utilize cross-lingual document representations.




  

\section{Related Works}

Crawling and mining the web for parallel data has been previously explored by~\citet{resnik1999mining} where the focus is on identifying parallel text from multilingual data obtained from a single source. For example, parallel corpora were curated from the United Nations General Assembly Resolutions~\cite{rafalovitch2009united,ziemski2016united} and from the European Parliament~\cite{koehn2005europarl}. 
However, curating from homogeneous sources by deriving domain-specific rules does not generalize to arbitrary web-domains.

Other approaches rely on metadata for mining parallel documents in unstructured web corpora. Some methods leveraged publication date and other temporal heuristics to identifying parallel documents~\cite{munteanu2005improving,munteanu2006extracting,udupa2009mint,do2009mining,abdui2009use}. However, temporal features are often sparse, noisy, and unreliable. Another class of alignment methods rely on document structure~\cite{resnik2003web,chen2000parallel} yet these structure signals can be sparse and may not generalize to new domains.

In the WMT-2016 bilingual document alignment shared task~\cite{buck2016findings}, many techniques were proposed to retrieve, score, and align cross-lingual document pairs. However this shared task only considered English to French -- a high-resource direction and the proposed techniques were not readily extendable to more languages.

Several approaches translate the target corpus into the source language, then apply retrieval and matching approaches on translated 2-grams and 5-grams to query, retrieve, and align documents~\cite{dara2016yoda,gomes2016first}. These methods rely on high-quality translation systems to translate, however such models may not exist, especially for low-resource language directions. Additionally, these methods leverage rare n-grams to identify likely candidates, yet low-frequency words and phrases that are likely to be mistranslated by machine translation systems.

In the shared task, many document similarity measures were investigated for use in aligning English to French web documents. One method utilized a phrase table from a phrase-based statistical machine translation system to compute coverage scores, based on the ratio of phrase pairs covered by a document pair~\cite{gomes2016first}. Other methods utilize the translated content of the target (French) document, and find the source (English) corresponding document based on n-gram matches in conjunction with a heuristic document length ratio~\cite{dara2016yoda,shchukin2016word}. Other methods translate the target documents into the source language and apply cosine similarity between tf/idf weighted vectors on unigrams and n-grams~\cite{buck2016quick,medvevd2016english,jakubina2016bad}. Finally, several methods were introduced that score pairs using metadata in each document such as links to documents, URLs, digits, and HTML structure~\cite{espla2016bitextor,papavassiliou2016ilsp}.

Recently, the use of neural embedding methods has been explored for bilingual alignment of text at the sentence and document level. One method proposes using hierarchical document embeddings, constructed from sentence embeddings, for bilingual document alignment~\cite{guo-EtAl:2019:WMT2}. Another method leverages a multilingual sentence encoder to embed individual sentences from each document, then performs a simple vector average across all sentence embeddings to form a dense document representation with cosine similarity guiding document alignment~\cite{el2019massive}.

Word mover's distance (WMD) is an adaptation of earth mover's distance (EMD)~\cite{rubner1998metric} that has been recently used for document similarity and classification~\cite{kusner2015word,huang2016supervised,atasu2017linear}. Other methods have leveraged the distance for cross-lingual document retrieval~\cite{balikas2018cross}. However these methods treat individual words as the base semantic unit for comparison which are intractable for large web-document alignment.

Finally, sentence mover's similarity has been proposed for automatically evaluating machine-generated texts outperforming ROUGE~\cite{clark2019sentence}. This method is purely monolingual and sentence representations are constructed by summing individual word embeddings.
\section{Problem Definition}
Given a set of source documents, $D_s$ and a set of target documents $D_t$, there exist $|D_s|\times |D_t|$ potential pairs of documents of the form $(d_s, d_t)$. Let $\mathcal{P}$ be the set of all candidate pairs ($D_s \times D_t$). Then cross-lingual document alignment aims to find the largest mapping from source documents to target documents, $\mathcal{P'} \subset \mathcal{P}$, s.t. given an $D_s$ and $D_t$ where, without a loss of generality, $|D_s| \leq |D_t|$, the largest \textit{injective function mapping} between $D_s$ and $D_t$:
\begin{equation*}
\forall a,b \in D_s, (a, c) \in \mathcal{P'} \land (b, c) \in \mathcal{P'}  \implies a=b 
\end{equation*}
In other words, each source document and target document can only be used in at most a single pair. This can be seen in Figure~\ref{fig:aligned_docs} where within the same web-domain, given source and target documents, the task is to match each source document to a unique target document where possible.

To find the best possible mapping between $D_s$ and $D_t$ we require two components: 1) a similarity function $\phi(d_s, d_t)$ which is used to score a set of candidate document pairs according to their semantic relatedness; and 2) an alignment or matching algorithm which uses the scores for each of the pairs in $D_s \times D_t$ to produce an alignment of size $min(|D_s|, |D_t|)$ representing the best mapping according to $\phi(d_s, d_t)$.




\section{Cross-Lingual Sentence Mover's Distance}
\label{sec:smd}

WMD fails to generalize to our use case for two reasons: (1) it relies on monolingual word representations which fail to capture the semantic distances between different language documents (2) intractability due to long web documents or lack word boundaries in certain languages.

To address this, we introduce cross-lingual sentence mover's distance (SMD) and show that representing each document as a bag-of-sentences (BOS) and leveraging recent improvements in multilingual sentence representations, SMD can better identify cross-lingual document pairs.
\subsection{Cross-Lingual Sentence Mover's Distance}
\label{sec:SMD}

Our proposed SMD solves the same optimization problem as WMD, but utilizes cross-lingual sentence embeddings instead of word embeddings as the base semantic. In particular, we utilize LASER sentence representations~\cite{artetxe2019massively}. LASER learns to simultaneously embed 93 languages covering 23 different alphabets into a joint embedding space by training a sequence-to-sequence system on many language pairs at once using a shared encoder and a shared byte-pair encoding (BPE) vocabulary for all languages. Utilizing LASER, each sentence is encoded using an LSTM encoder into a fixed-length dense representation. 

We adapt EMD to measure the distance between two documents by comparing the distributions of sentences within each document. More specifically, SMD represents each document as a \emph{normalized bag-of-sentences} (nBOS) where each sentence has associated with it some probability mass. As distances can be computed between dense sentence embeddings, the overall document distance can then be computed by examining how close the distribution of sentences in the source document is to sentences in the target document. We formulate this distance as the minimum cost of transforming one document into the other. 

For our basic formulation of SMD, each document is represented by the relative frequencies of sentences, i.e., for the $i_{th}$ sentence in the document,

\begin{equation}
\label{eq:sent_weighting}
d_{A,i} = cnt(i)/ |A|
\end{equation}

where $|A|$ is the total number of sentence in document A, and d$_{B,i}$ is defined similarly for document B. Under this assumption, each individual sentence in a document is equally important and probability mass is allocated uniformly to each sentence. Later, we will investigate alternative schemes to allocating probability mass to sentences.

Now let the $i_{th}$ sentence be represented by a vector $v_i \in R^m$. This length-m dense embedding representation for each sentence allows us to define distances between the i$_{th}$ and j$_{th}$ sentences. We denote $\Delta(i,j)$ as the distance between the $i_{th}$ and $j_{th}$ sentences and let $V$ denote the vocabulary size where the vocabulary is the unique set of sentences within a document pair. We follow previous works~\cite{kusner2015word} and use the Euclidean distance, $\Delta(i,j) = ||v_i - v_j||$. The SMD between a document pair is then the solution to the linear program:

\begin{equation}
    SMD(A,B) = \underset{T\ge0}{min} \sum_{i=1}^V\sum_{j=1}^V T_{i,j} \times \Delta(i,j)
\end{equation}

subject to:  
\begin{align*}
    &\forall i \sum_{j=1}^V T_{i,j} = d_{A,i} \\
    &\forall j \sum_{i=1}^V T_{i,j} = d_{B,j}
\end{align*}

Where $T\in R^{V\times V}$ is a nonnegative matrix, where each $T_{i,j}$ denotes how much of sentence $i$ in document $A$ is assigned to sentences $j$ in document $B$, and constraints ensure the flow of a given sentence cannot exceed its allocated mass. Specifically, SMD ensures the the entire outgoing flow from sentence $i$ equals $d_{A,i}$, i.e. $\sum_{j} T_{i,j} = d_{A,i}$. Additionally, the amount of incoming flow to sentence $j$ must match $d_{B,j}$, i.e., $\sum_{i} T_{i,j} = d_{B,j}$.

\subsection{Alternative Sentence Weighting Schemes}
\label{sec:alternative}
In Equation~\ref{eq:sent_weighting}, each document is represented as a normalized bag-of-sentences (nBOS) where sentences are equally weighted. However, we posit that some sentences may be more semantically important than others.

\paragraph{\textbf{Sentence Length Weighting}}
The first insight we investigate is that documents will naturally be segmented into sentences of different lengths based on the language, content, and choice of segmentation. While Equation~\ref{eq:sent_weighting}, treats each sentence equally, we posit that longer sentences should be assigned larger weighting than shorter sentences. 

As such, we weight each sentence by the number of tokens in the sentence relative to the total number of tokens in the entire document, i.e., for the $i_{th}$ sentence in the document $A$, we compute the weighting SL(i) as follows:

\begin{equation}
d_{A,i} = cnt(i) \cdot |i|/ \underset{s\in A}{\sum} cnt(s)\cdot |s|
\end{equation}

where $|i|$ and $|s|$ indicate the number of tokens in sentences $i$ and $s$ respectively. As such, longer sentence receive larger probability mass than shorter sentences.

\paragraph{\textbf{IDF Weighting}}
The second insight we investigate is that text segments such as titles and navigation text is ubiquitous in crawled data yet less semantically informative. Based on this insight, we apply a variant of inverse document frequency (IDF) -- a weighting scheme common in the information retrieval space -- to individual sentences~\cite{robertson2004understanding}. Under this scheme, the more common a sentence is within a webdomain, the less mass the sentence will be allocated.

For sentence $i$ in a web-domain $D$, we compute IDF(i) as follows:
\begin{equation}
d_{A,i} = 1 + \log \frac{|D|}{|\{d \in D: i \in d\}|}
\end{equation}

where  $|\{d \in D: s \in d\}|$ is the number of documents where the sentence $s$ occurs and smoothing by $1$ is performed to prevent 0 IDF. 

\paragraph{\textbf{SLIDF Weighting}}
Finally, we propose combining both sentence length and inverse document frequency into a joint weighting scheme:

\begin{equation}
d_{A,i} = SL(i) \cdot IDF(i)
\end{equation}

In this scheme, each sentence is weighted proportionally to the number of tokens it contains as well as by the IDF of the sentence within the domain. This weighting scheme is reminiscent of the use of tf-idf to determine word relevance~\cite{ramos2003using}, but instead sentence length and idf are used to determine sentence importance.

\subsection{Fast Distance Approximation}
\label{sec:fast_approximation}
 While EMD and other variants have demonstrated superior performance in many retrieval and classification tasks, they have also been shown to suffer from high computational complexity $\mathcal{O}(p^3\log p)$, where $p$ denotes the number of unique semantic units in a document pair. As such, we investigate techniques to speed up this computation.

\paragraph{\textbf{Relaxed SMD}}
Given the scalability challenges for computing WMD, simplified version of WMD was proposed that relaxes one of the two constraints in the original formulation~\cite{kusner2015word}. Applying the same principle to SMD, we formulate: 

\begin{equation*}
    SMD(A,B) = \underset{T\ge0}{min} \sum_{i=1}^V\sum_{j=1}^V T_{i,j} \times \Delta(i,j)
\end{equation*}

subject to: $\forall i \sum_{j=1}^V T_{i,j} = d_{A,i}$.
Analogous to the relaxed-WMD, this relaxed problem yields a lower-bound to the SMD as every SMD solution satisfying both constraints remains a feasible solution if one constraint is removed. The optimal solution can be found by simply allocating the mass in each source sentence to the closest sentence in the target document.

The same computation can be performed in the reverse direction by removing the second constraint: $\forall j \sum_{i=1}^V T_{i,j} = d_{B,j}$. Similarly, the optimal solution allocates the mass sentences in the target document to the closest sentence in the source document. Both these distances can be calculated by computing the distance matrix between all pairs of sentences in $\mathcal{O}(p^2)$ time. For a tighter estimate of distance, the maximum of the two resultant distances can be used.

\paragraph{\textbf{Greedy Mover's Distance}}
We introduce an alternative to the relaxed-EMD variant wherein we keep both constraints in the transportation problem, but identify an approximate transportation scheme. This greedy mover's distance (GMD)  finds the closest sentence pair between the source and target and moves as much mass between the two sentences as possible; the algorithm moves to the next closest until all mass has been moved while maintaining both constraints.

\begin{algorithm2e}[ht]
\caption{Greedy Mover's Distance}
\label{alg:greedy_distance}
\Indm
\small
       \KwIn{$d_s, d_t, w_s, w_t$} 
       \KwOut{$\Delta(d_s, d_t)$}
\Indp
       \BlankLine
    $pairs$ $\gets \{(s_s, s_t) \text{ for } s_s, s_t  \in d_s \times d_t\}$ \text{ in ascending order by }   $\lVert s_s - s_t\rVert$ \\
	 $\text{distance } \gets 0.0$ \\
	\For{$s_s, s_t \in$ pairs}{
		   flow $\gets min(w_s[s_s], w_t[s_t])$ \\
		   $w_s[s_s] \gets w_s[s_s] - \text{flow}$ \\
		   $w_t[s_t] \gets w_t[s_t] - \text{flow}$ \\
		   $\text{distance} \gets \text{distance} + \lVert s_s - s_t\rVert \times \text{flow}$\\
			
	}	
	  \textbf{return} total
\end{algorithm2e}
\normalsize

As seen in Algorithm~\ref{alg:greedy_distance}, the algorithm takes a source document ($d_s$) and a target document ($d_t$) as well as the probability mass for the sentences in each: respectively $w_s$ and $w_t$. The algorithm first computes the euclidean distance between each sentence pair from source to target and sorts these pairs in ascending order by their euclidean distance. The algorithm then iteratively chooses the closest sentence pair and moves the mass of the smallest sentence from the source to the target and subtracting this moved math from both. The algorithm terminates when all moveable mass has been moved. Unlike the exact solution to EMD, the runtime complexity is a more tractable $\mathcal{O}(|d_s||d_t|\times\log(|d_s||d_t|))$ which is dominated by the cost of sorting all candidate pairs. Unlike the relaxation, both constraints are satisfied but the transport is not necessarily optimal. As such, GMD yields an upper-bound to the exact computation. 

We experimentally compare the effect of both approximation strategies on downstream document alignment in Section~\ref{sec:discussion}.

\section{Document Matching Algorithm}
\label{sec:alignment}
In addition to a distance metric (i.e. SMD), we need a document matching algorithm to determine the best mapping between documents in two languages. 

In our case, this works as follows: for any given webdomain, each document in the source document set, $D_s$ is paired with each document in the target set, $D_t$, yielding $|D_s \times D_t|$ scored pairs -- a fully connected bipartite graph representing all candidate pairings. Similar to previous works~\cite{buck2016quick}, the expected output assumes that each webpage in the non-dominant language has a translated or comparable counterpart. As visualized in Figure~\ref{fig:aligned_docs}, this yields a $min(|D_s|, |D_t|)$ expected number of aligned pairs.

While an optimal matching maximizing scoring can be solved using the Hungarian algorithm~\cite{munkres1957algorithms}, the complexity of this algorithm is $\mathcal{O}(max(|D_s||D_t|)^3)$ which is intractable to even moderately sized web domains. As such, similar to the work in~\cite{buck2016quick}, a one-to-one matching between English and non-English documents is enforced by applying, competitive matching, a greedy bipartite matching algorithm. 

\begin{algorithm2e}[ht]
\caption{Competitive Matching}
\label{alg:greedy}
\Indm
\small
      \KwIn{$P = \{(d_s, d_t) | d_s \in D_s, d_t \in D_t\}$} 
      \KwOut{$P'= \{(d_{s,i},d_{t,i}), . . .\} \subset P$}
\Indp
      \BlankLine
    $scored$ $\gets \{(p, score(p)) \text{ for } p \in P\}$ \\
	$sorted$ $\gets sort(scored) \text{ in ascending order } $\\
	aligned $\gets \varnothing$\\
	$S_s \gets \varnothing$\\
	$S_t \gets \varnothing$ \\
	\For{$d_s, d_t \in$ sorted}{
	    \uIf{$d_s \notin S_s \land d_t \notin S_t$}  
		{
		   $aligned \gets aligned \cup \{(d_s, d_t)\}$ \\
		   $S_s \gets S_s \cup d_s$ \\
		   $S_t \gets S_t \cup d_t$ \\
		}	
	}	
	  \textbf{return} aligned
\end{algorithm2e}
\normalsize

In Algorithm~\ref{alg:greedy}, the algorithm first scores each candidate document pair using a distance function and then sorts pairs from closest to farthest. The algorithm then iteratively selects the closest document pair as long as the $d_s$ and $d_t$ of each pair have not been used in a previous (closer) pair. The algorithm terminates when $min(|D_s|, |D_t|)$ pairs have been selected. Unlike the Hungarian algorithm, the runtime complexity is a more tractable $\mathcal{O}(|D_s||D_t|\times\log(|D_s||D_t|))$ which is dominated by the cost of sorting all candidate pairs.

\begin{table*}[t]
\begin{minipage}[b]{0.28\linewidth}\centering
    \scriptsize
    \subcaptionbox{High-resource directions. \label{subtable:highresource_baseline}}{
    \setlength\tabcolsep{2.0pt}
        \begin{tabular}{l@{\hspace{4pt}}  c@{\hspace{4pt}}  c@{\hspace{4pt}}  c@{\hspace{4pt}}  c@{\hspace{4pt}} c@{\hspace{4pt}} >{\bf} c@{\hspace{4pt}}}
            \toprule
            & \multicolumn{6}{c}{{\bf Recall}}\\\cmidrule{2-7}
            \textbf{Language} & \textbf{DE} & \textbf{SA}  & \textbf{SMD} & \textbf{SL} & \textbf{IDF}& \textbf{SLIDF}\\\midrule
            French & 0.39 & 0.84  & 0.81 & 0.84 & 0.83 & 0.85\\
            Spanish & 0.34 & 0.53   & 0.59 & 0.63 & 0.62 & 0.64\\
            Russian  & 0.06 & 0.64  & 0.69 & 0.69 & 0.70 & 0.71\\
            German & 0.52 & 0.74   & \textbf{0.78} & 0.76 & 0.77 & \textnormal{0.77}\\
            Italian & 0.22 & 0.47   & 0.55 & 0.56 & 0.56 & 0.59\\
            Portuguese  & 0.17 & 0.36   & 0.39 & \textbf{0.41} & 0.38 & \textnormal{0.40}\\
            Dutch & 0.28 & 0.49   & 0.54 & 0.54 & 0.54 & 0.56\\
            Indonesian  & 0.11 & 0.47   & 0.49 & 0.52 & 0.51 & 0.53\\
            Polish  & 0.17 & 0.38   & 0.45 & 0.45 & \textbf{0.46} & 0.46\\
            Turkish  & 0.12 & 0.38   & 0.52 & 0.56 & 0.57 & 0.59\\
            Swedish  & 0.19 & 0.40   & 0.44 & 0.44 & \textbf{0.46} & \textnormal{0.45} \\
            Danish  & 0.27 & 0.62   & 0.63 & \textbf{0.69} & 0.65 & 0.69\\
            Czech  & 0.15 & 0.40   & 0.43 & \textbf{0.44} & \textbf{0.44} & \textnormal{0.43}\\
            Bulgarian  & 0.07 & 0.43  & 0.52 & 0.54 & \textbf{0.55} & \textnormal{0.52}\\
            Finnish  & 0.06 & 0.47  & 0.51 & 0.51 & \textbf{0.54} & \textnormal{0.52}\\
            Norwegian  & 0.13 & 0.33  & 0.37 & 0.39 & \textbf{0.42} & \textnormal{0.41}\\
            \midrule 
            \textbf{AVG} & 0.20 & 0.50  & 0.54 & 0.56 & 0.56 & 0.57\\
            \bottomrule 
        \end{tabular}
    }
    \end{minipage}
    \hspace{0.5cm}
    \begin{minipage}[b]{0.28\linewidth}
    \scriptsize
    \centering
        \subcaptionbox{Mid-resource directions. \label{subtable:mindresource_baseline}}{
        \setlength\tabcolsep{2.0pt} 
        \begin{tabular}{l@{\hspace{4pt}}  c@{\hspace{4pt}}  c@{\hspace{4pt}}  c@{\hspace{4pt}}  c@{\hspace{4pt}} c@{\hspace{4pt}} >{\bf} c@{\hspace{4pt}}}
            \toprule
            & \multicolumn{6}{c}{{\bf Recall}}\\\cmidrule{2-7}
            \textbf{Language} & \textbf{DE} & \textbf{SA} & \textbf{SMD} & \textbf{SL} & \textbf{IDF}& \textbf{SLIDF}\\\midrule
            Romanian  &  0.15 & 0.40  & 0.44  & 0.43 & \textbf{0.45} & \textnormal{0.43}\\
            Vietnamese  &  0.06 & 0.28 & 0.29  & 0.29 & 0.29  & 0.32\\
            Ukrainian  & 0.05 & 0.68  & 0.67  & 0.78 & 0.78 & 0.82\\
            Greek  & 0.05 & 0.31  & 0.47  & 0.48 & \textbf{0.49}  & 0.49\\
            Korean  & 0.06 & 0.34  & 0.60  & 0.54 & \textbf{0.61} & \textnormal{0.60}\\
            Arabic  & 0.04 & 0.32  & 0.63  & 0.59 & \textbf{0.65} & \textnormal{0.61}\\
            Croatian  & 0.16 & 0.37  & 0.40  & 0.40  & \textbf{0.41} & \textnormal{0.40}\\
            Slovak  & 0.20 & 0.41  & 0.46  & \textbf{0.46} & \textbf{0.46} & \textnormal{0.44}\\
            Thai  & 0.02 & 0.19  & 0.41  & 0.33 & \textbf{0.47} & \textnormal{0.41}\\
            Hebrew  & 0.05 & 0.18  & 0.39  & \textbf{0.43} & 0.41 & \textnormal{0.41}\\
            Hindi  & 0.04 & 0.27  & 0.34  & \textbf{0.54} & \textnormal{0.52} & \textnormal{0.53}\\
            Hungarian  & 0.15 & 0.49  & 0.50  & \textbf{0.54}  & 0.51 & 0.54\\
            Lithuanian  & 0.11 & 0.73  & 0.79  & 0.79 & \textbf{0.80} & 0.80\\

            Slovenian  & 0.13 & 0.33  & 0.34  & 0.35 & \textbf{0.36} & 0.36\\
            Persian  & 0.06 & 0.32 & 0.56  & 0.57 & 0.53 & 0.59\\
            \\
            
           \midrule 
           \textbf{AVG} & 0.09 & 0.37  & 0.49  & 0.50 & \textbf{0.52} & 0.52\\
            \bottomrule 
        \end{tabular}
    }
    \end{minipage}
        \hspace{0.5cm}
    \begin{minipage}[b]{0.28\linewidth}\centering
    \scriptsize
    \subcaptionbox{Low-resource directions. \label{subtable:lowresource_baseline}}{
        \setlength\tabcolsep{2.0pt}
        \begin{tabular}{l@{\hspace{4pt}}  c@{\hspace{4pt}}  c@{\hspace{4pt}}  c@{\hspace{4pt}}  c@{\hspace{4pt}} c@{\hspace{4pt}} >{\bf} c@{\hspace{4pt}}}
            \toprule
            & \multicolumn{6}{c}{{\bf Recall}}\\\cmidrule{2-7}
            \textbf{Language} & \textbf{DE} & \textbf{SA} & \textbf{SMD} & \textbf{SL} & \textbf{IDF} & \textbf{SLIDF} \\\midrule
            
            Estonian  & 0.28 & 0.52 & 0.69 &  0.66 &  \textbf{0.74} & \textnormal{0.72}\\

            Bengali & 0.05 & 0.32  & 0.78 &  0.72 &  0.77 & 0.79\\
            Albanian  & 0.23 & 0.56 & \textbf{0.66} &  0.65 &  0.65 & 0.66\\
            Macedonian  & 0.02 & 0.33 & 0.32 &  0.36 &  \textbf{0.38} & \textnormal{0.33}\\
            Urdu  & 0.06 & 0.22 & \textbf{0.60} &  \textbf{0.60} &  0.49 & \textnormal{0.56}\\
            Serbian  & 0.06 & 0.59  & \textbf{0.75} &  0.74 &  0.74 & \textnormal{0.71}\\
            Azerbaijani	&  0.08  & 0.34  & 0.74 & 0.74 &  \textbf{0.75} & \textnormal{0.74}\\
            Armenian  & 0.02 & 0.18 & 0.32 &  0.35 &  0.34 & 0.38\\
            Belarusian  & 0.07 & 0.47 & 0.67 &  0.69 &  \textbf{0.73} & \textnormal{0.71}\\
            Georgian  & 0.06 & 0.24  & 0.46 &  \textbf{0.48} &  0.45 & \textnormal{0.45}\\
            Tamil  & 0.02 & 0.20  & 0.51 &  0.45 &  0.51 & 0.53\\
            Marathi  & 0.02 & 0.11  & 0.43 &  \textbf{0.46} &  0.33 & \textnormal{0.39}\\
            Kazakh  & 0.05 & 0.31  & 0.44 &  \textbf{0.46}  &  0.45 & \textnormal{0.45}\\
            Mongolian  & 0.03 & 0.13  & 0.18 &  0.22 &  0.21 & 0.23\\
            Burmese  & 0.01 & 0.10  & 0.26 &  0.33 &  \textbf{0.46} & 0.46\\
            Bosnian  & 0.18 & 0.64 & 0.61 &  0.69  &  0.65 & 0.72\\
            
            \midrule 
            \textbf{AVG} & 0.08 & 0.33  & 0.53 &  0.54 &  0.54  & 0.55\\
            \bottomrule 
        \end{tabular}
        }

    \end{minipage}
    \caption{Alignment recall on URL-aligned CommonCrawl dataset.} 
    \label{tab:results_baselines}
\end{table*}
\normalsize

\section{Experiments and Results}
\label{sec:results}
In this section, we explore the question of whether SMD can be used as a dissimilarity metric for the document alignment problem. Moreover, we explore which sentence weighting schemes yield the best results.

\subsection{Experimental Setup}
\paragraph{\textbf{Dataset}}
We evaluate on the test set from the URL-Aligned CommonCrawl dataset~\cite{el2019massive} across 47 language directions.

\paragraph{\textbf{Baseline Methods}}
For comparison, we implemented two existing and intuitive document scoring baselines from~\cite{el2019massive}. The direct embedding (DE), directly embeds the entire content of a document using LASER. The second method sentence averaging (SA) embeds all sentences in a document using LASER and averages all embeddings to get a document representation. Cosine similarity on the embedded representation is used to compare documents. 

\paragraph{\textbf{SMD Weightings}}
We evaluate four weighting schemes for SMD: (1) vanilla SMD with each sentence equally weighted(2) weighting by sentence length (SL) where SMD is computed under a scheme where each sentence is weighted by its length (number of tokens) normalized by the length of the entire document (3) weighting by inverse document frequence (IDF) where SMD is computed under a scheme where each sentence is weighted by the idf of the sentence (4) computing SMD under a scheme where each sentence is weighted by both sentence length and inverse document frequency (SLIDF). Under all these schemes, all weights are normalized to unit measure.

\paragraph{\textbf{Distance approximation}}
We use the greedy mover's distance approximation for all variants reported. In Section~\ref{sec:discussion} we further explore the performance of the full distance computation and relaxed variants that were described in Section~\ref{sec:fast_approximation}.

\paragraph{\textbf{Evaluation Metric for Document Alignment}}
Because the ground-truth document pairs only reflect a high-precision set of web-document pairs that are translations or of comparable content, there may be many other valid cross-lingual document pairs within each web-domain that are not included in the ground truth set. As such, we evaluate each method's generated document pairs solely on the recall (i.e. what percentage of the aligned pages in the test set are found) from the ground truth pairs.

For each scoring method, we score document pairs from the source and target languages within the same web-domain using the proposed document distance metrics described above. For the alignment, we report the performance for each distance metric after applying the competitive matching alignment algorithm as described in Algorithm~\ref{alg:greedy}. 





\subsection{Results}
\label{sec:alignment_experiments}
In Table~\ref{tab:results_baselines}, we first notice that constructing document representations by directly embedding (DE) the entire content of each document and computing document similarity using cosine similarity of the representation severely under-performs compared to individually embedding sentences and constructing the document representations by averaging the individual sentence representations within the document (SA). This is intuitive as LASER embeddings were trained on parallel sentences and embedding much larger documents directly using LASER results in poorer representations than by first embedding smaller sentences and combining them into the final document representation. 

Comparing the basic SMD to the best performing baseline (SA), we see a 4\%, 12\%, and 20\% improvement across high, mid, and low-resource directions respectively. This improvement suggests that summing sentence embeddings into a single document representation degrades the quality of the resultant document distances over computing document distances by keeping all sentence representations separate and computing distances between individual sentence pairs and combining these distances into a final document distance. This is more pronounced in lower-resource over higher-resource pairs which may be due to poorer lower-resource embeddings due to LASER being trained on fewer low-resource sentence pairs. As such averaging is more destructive to these representations while SMD avoids this degradation.

Further analysis verified the intuition that different sentences should be allocated different weighting in SMD. Assigning mass proportional to the number of tokens in the sentence (SL), we see a 2\%, 1\% and 1\% absolute improvement in recall in high, mid, and low-resource directions over assigning equal probability mass. This supports the claim that longer sentences should be allocated higher importance weight over shorter sentences as they contain more semantic content. The second assumption we investigated is that sentences that are common within a webdomain have less semantic importance and should be allocated less probability mass when computing SMD. After computing SMD with each sentence allocated mass according to inverse document frequency (IDF) and normalized to unit measure, we see a 2\%, 3\%, and 1\% improvement over SMD for high, mid, and low-resource directions. Finally, when combining both sentence length and inverse document frequency (SLIDF) and normalizing to unit measure, we see a 3\%, 3\% and 2\% absolute improvement in recall for high, mid, and low-resource directions.  Overall, our SMD with SLIDF weighting scheme outperforms the sentence averaging baseline by 7\% on high-resource directions, 15\% on mid-resource directions, and 22\% on low-resource directions.

\section{Discussion}
\label{sec:discussion}

Although using sentences over words as the base semantic unit drastically reduces the overall cost of computing EMD-based metrics, the cubic computation still prohibits its use as a fast distance metric for large-scale alignment efforts. As such, in Section~\ref{sec:fast_approximation} we described two faster approximations to EMD computation: (1) a relaxation of constraints resulting in a lower bound and (2) a greedy algorithm for computing assigning transport representing an upper bound. We first analyze and compare the distances from each approximation scheme to the exact SMD computation. 

\begin{figure*}[t]

  \begin{subfigure}[t]{5.0cm}
    \includegraphics[width=5.0cm]{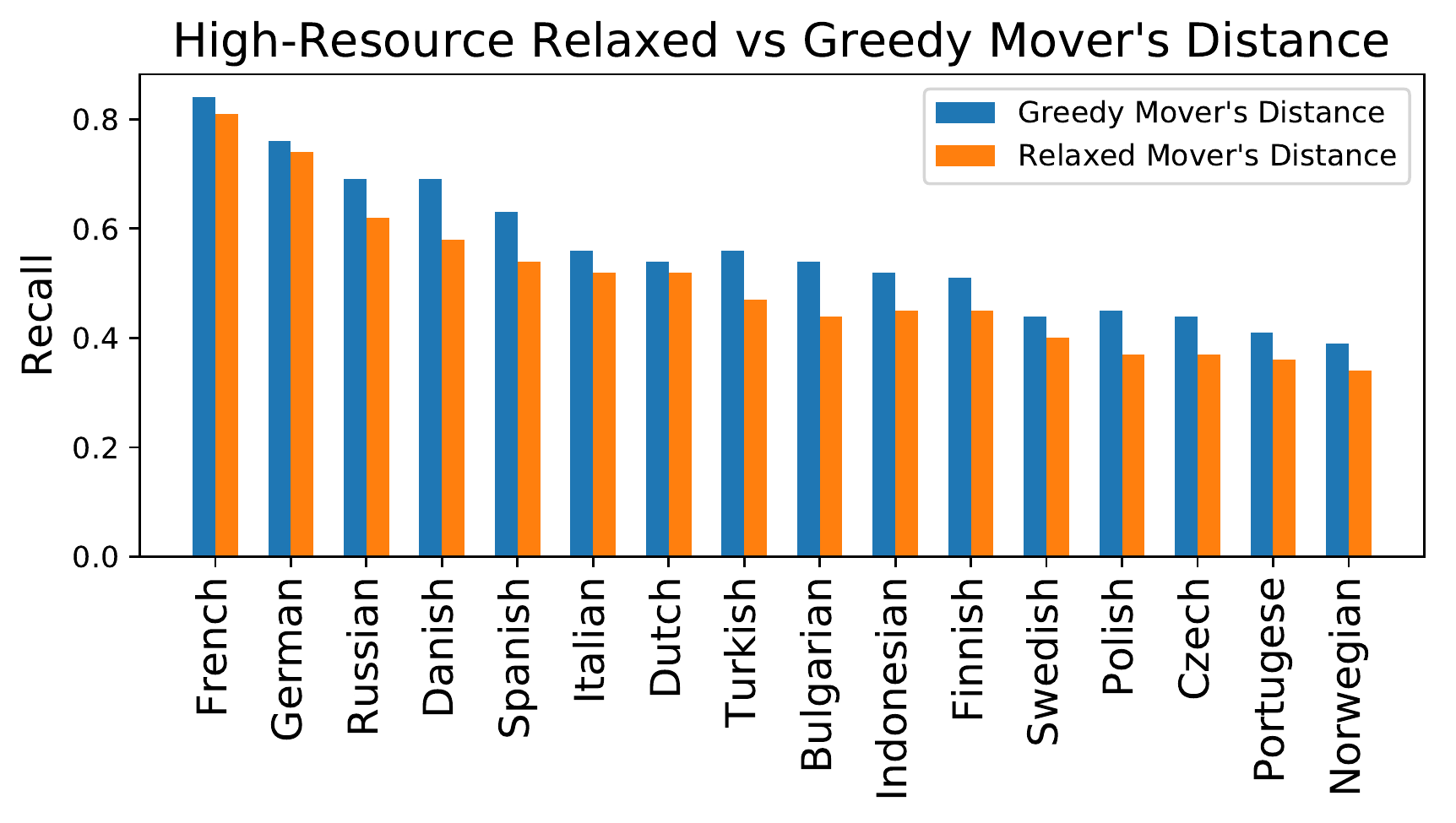}
    \caption{\footnotesize{High-resource directions.}}
  \end{subfigure}
  \begin{subfigure}[t]{5.0cm}
    \includegraphics[width=5.0cm]{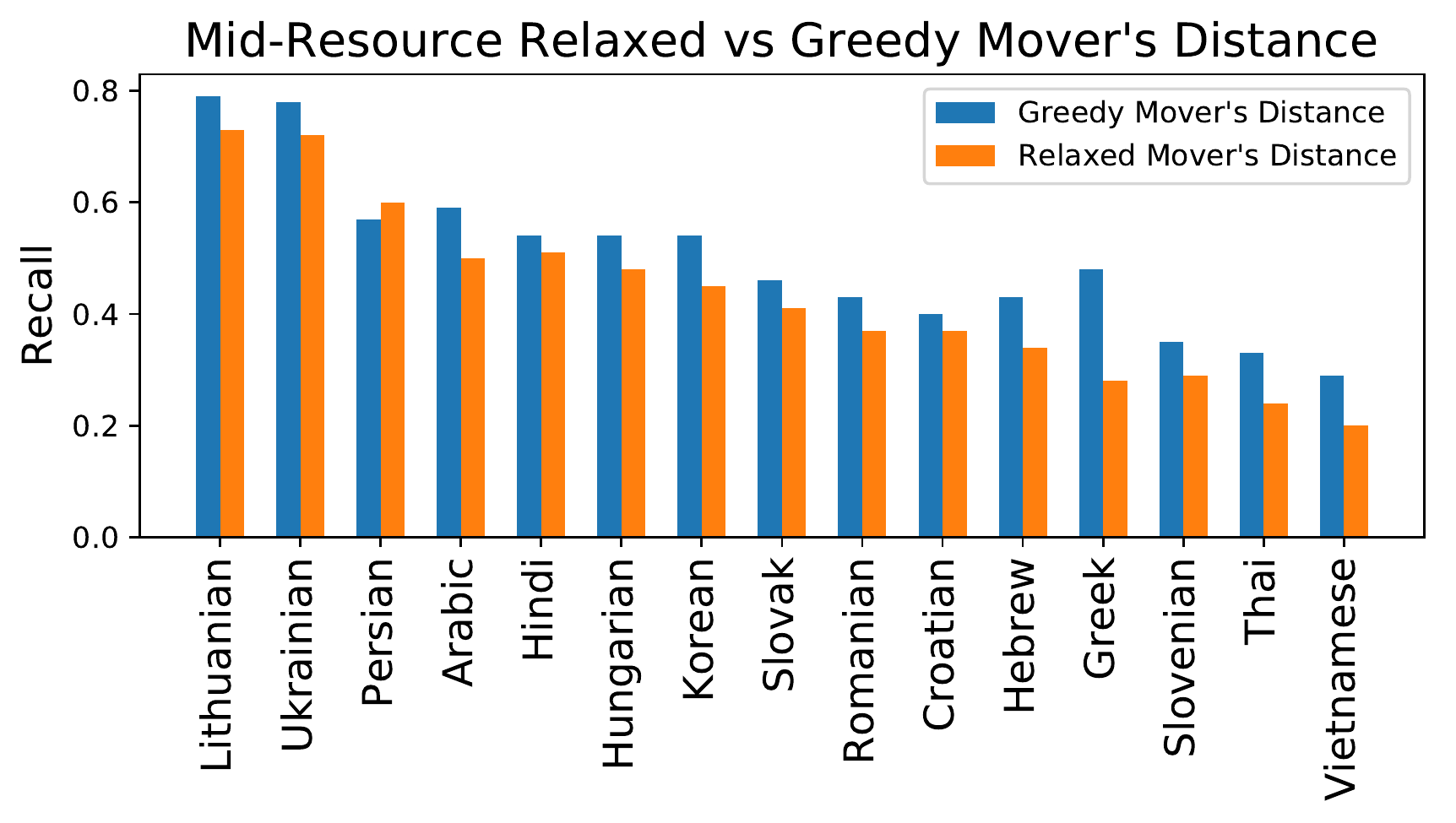}
    \caption{\footnotesize{Mid-resource directions.}}
  \end{subfigure}
  \begin{subfigure}[t]{5.0cm}
    \includegraphics[width=5.0cm]{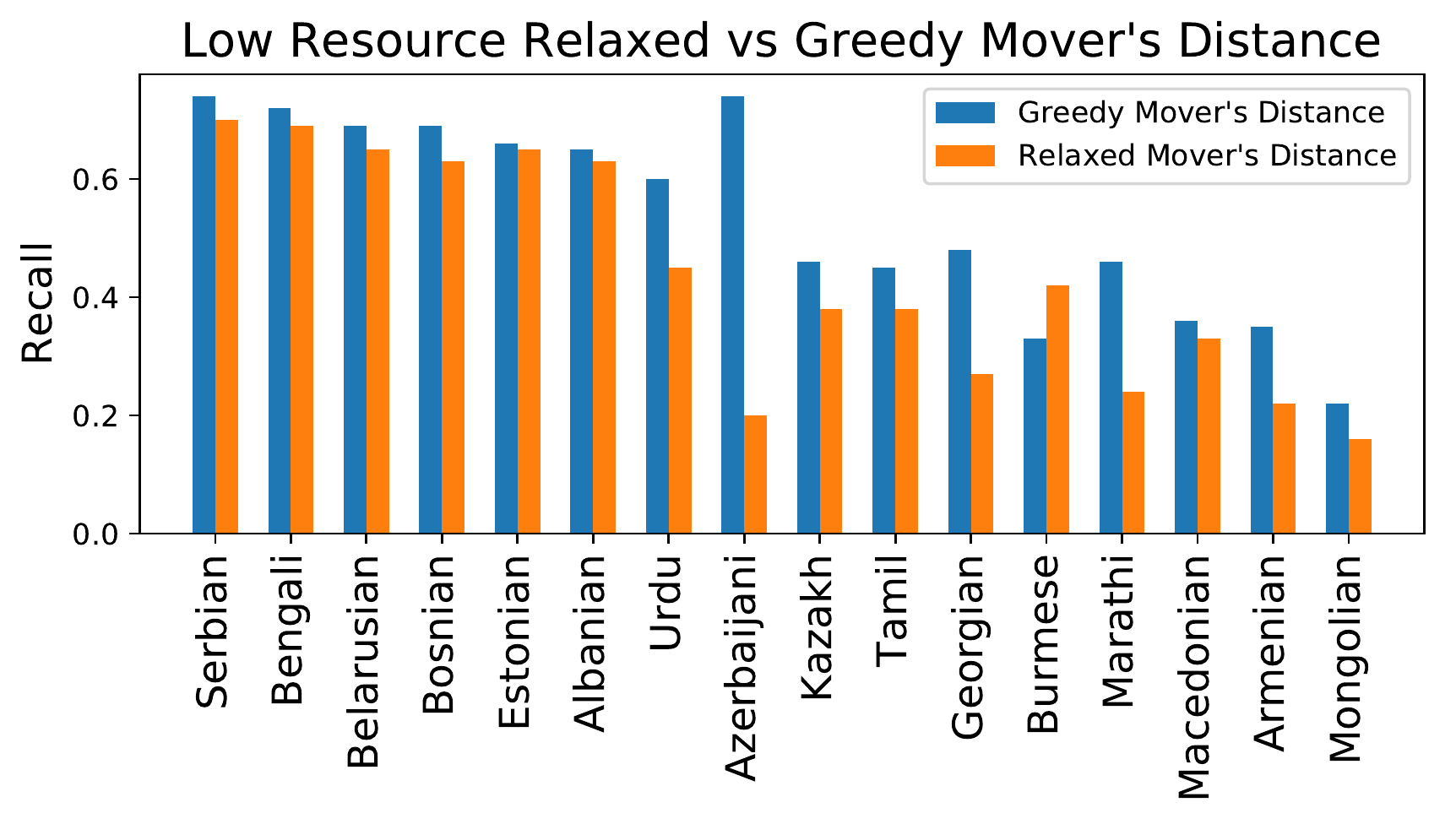}
    \caption{\footnotesize{Low-resource directions.}}
  \end{subfigure}
  \caption{Document alignment results for different distance approximation techniques.}
  \label{fig:approximations}
\end{figure*}

\begin{figure}[h]
    \centering
    \includegraphics[width=3.0in]{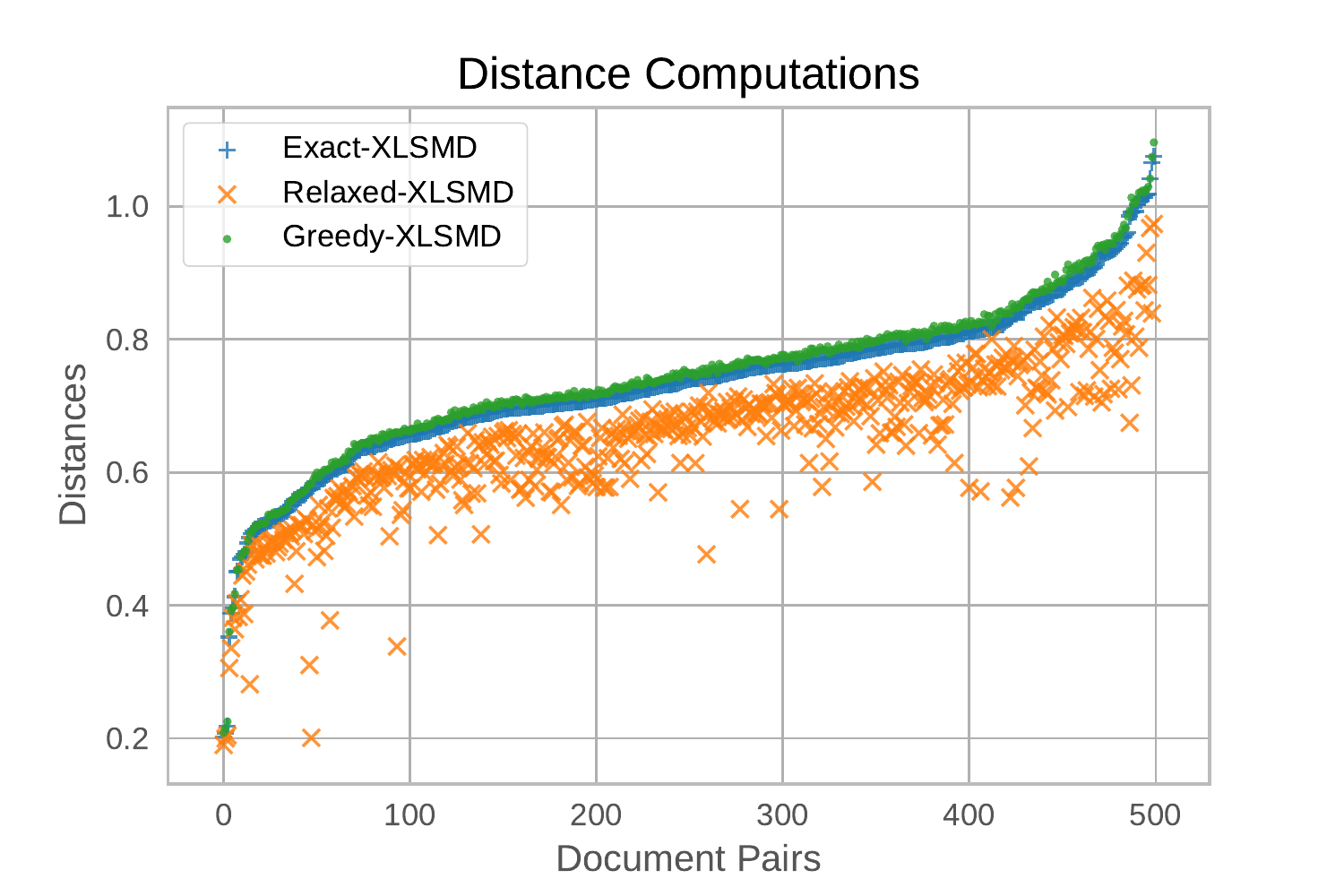}
    \caption{Exact, relaxed, and greedy-SMD distances sorted by Exact-SMD for a random selection of document pairs.}
    \label{fig:distance_comparison}
\end{figure}

\begin{table}[h]    \small
    \centering
    \begin{tabular}{l  c  c  c c}
        \toprule
        \textbf{Method} &  \textbf{Tau} & \textbf{Recall} & \textbf{MAE} & \textbf{Runtime (s)}  \\\midrule
        Exact-SMD & 1.00  & 0.69 & 0.000  & 0.402\\
        Relaxed-SMD & 0.70  & 0.58 & 0.084  & 0.031\\
        Greedy-SMD & 0.98  & 0.69 & 0.010 & 0.107\\\bottomrule
    \end{tabular}
    \caption{Comparing exact SMD computation to approximation schemes for computing SMD on 10 webdomains. }
    \label{tab:approx_vs_exact}
\end{table}
In Figure~\ref{fig:distance_comparison}, we see that the distance computations for exact SMD and the greedy SMD approximation are highly correlated with small variance, while the relaxed approximation is less so with high variance. Additionally, as discussed in Section~\ref{sec:fast_approximation}, the visualizations empirically suggest that our greedy approximation is a fairly tight upper bound while the relaxed approximation is a looser lower bound.

In Table~\ref{tab:approx_vs_exact}, we compare quantitative metrics for the relaxed and greedy approximations to the exact solution of SMD on ten webdomains. Our first evaluation investigates how the approximate computation of distances affects the resultant ordering of document pairs. For the ten selected webdomains, we sort the document pairs in order by their computed distances and compare the ordering to the ordering induced by the exact computation of SMD. We evaluate the orderings using the Kendall-Tau metric~\cite{kendall1938new} which measures the agreement between the two rankings; if the agreement between the two rankings is perfect (i.e.,  the two rankings are the same) the coefficient has value 1 and if the disagreement between the two rankings is perfect (i.e.,  one ranking is the reverse of the other) the coefficient has value -1. Intuitively, we would like the distances computed by an approximation to induce a similar ordering to the ordering by the exact distance computation. Comparing the Kendall-Tau for the relaxed and greedy approximations in relation to the exact computation shows that the order induced by the greedy approximation is very similar to the ordering induced by the exact computation while the relaxed approximation varies considerably.  Additionally, the relaxed approximation demonstrates fairly high mean absolute error (MAE) and results in lower document alignment recall when compared to the exact computation of SMD, while our greedy approximation performs comparably and shows insignificant MAE. Finally, while the runtime of the relaxed computation is the fastest at 13 times faster than the exact computation, our greedy algorithm is approximately 4 times faster while delivering comparable document alignment performance to the exact computation and superior performance to the relaxed computation.

To ensure that the greedy algorithm consistently outperforms the relaxed algorithm on document alignment, we investigate the effect of using each approximation method on the downstream document alignment performance across 47 language pairs of varying resource availability.
\begin{table}[h]    \small
    \centering
    \begin{tabular}{l  c  c  c  c}
        \toprule
        \textbf{Approximation} &  \textbf{Low} & \textbf{Mid} & \textbf{High}  & \textbf{All} \\\midrule
        Relaxed-SMD & 0.44  & 0.43 & 0.50 & 0.46 \\
        Greedy-SMD & 0.54  & 0.50 & 0.56 & 0.54\\\bottomrule 
    \end{tabular}
    \caption{Document alignment performance of fast methods for approximating the same variant of SMD.}
    \label{tab:approx}
\end{table}

As seen in Figure~\ref{fig:approximations}, in 45 of the 47 evaluated language pairs, our proposed Greedy Mover's Distance approximation yielded higher downstream recall in our alignment task over using the relaxed distance proposed for use in WMD~\cite{kusner2015word}. In Table~\ref{tab:approx}, we see a 10\%, 7\%, and 6\% improvement in downstream recall across low, mid, and high-resource directions respectively. These results indicate that relaxing one of the two constraints in EMD is too lax for measuring an accurate distance. We posit this is because there are many sentences that can be considered ``hubs" that are semantically close to many other sentences. These sentences can have a lot of probability mass allocated to them, resulting in a lower approximate EMD. Our greedy approximation ensures that both constraints are maintained even if the final result does not reflect the optimal transport. 

\section{Conclusion}
\label{sec:conclusion}
In this paper, we introduce SMD a cross-lingual sentence mover's distance metric for automatically assessing the semantic similarity of two documents in different languages. We leverage state-of-the-art multilingual sentence embeddings and apply SMD to the task of cross-lingual document alignment. We demonstrate that our new metric outperforms other unsupervised metrics by a margin, especially in medium and low-resourced conditions.

\balance

\bibliography{aacl-ijcnlp2020}

\begin{thebibliography}{37}
\expandafter\ifx\csname natexlab\endcsname\relax\def\natexlab#1{#1}\fi

\bibitem[{AbduI-Rauf and Schwenk(2009)}]{abdui2009use}
Sadaf AbduI-Rauf and Holger Schwenk. 2009.
\newblock On the use of comparable corpora to improve smt performance.
\newblock In \emph{Proceedings of the 12th Conference of the European Chapter
  of the Association for Computational Linguistics}, pages 16--23. Association
  for Computational Linguistics.

\bibitem[{Artetxe and Schwenk(2019)}]{artetxe2019massively}
Mikel Artetxe and Holger Schwenk. 2019.
\newblock Massively multilingual sentence embeddings for zero-shot
  cross-lingual transfer and beyond.
\newblock \emph{Transactions of the Association for Computational Linguistics},
  7:597--610.

\bibitem[{Atasu et~al.(2017)Atasu, Parnell, D{\"u}nner, Sifalakis, Pozidis,
  Vasileiadis, Vlachos, Berrospi, and Labbi}]{atasu2017linear}
Kubilay Atasu, Thomas Parnell, Celestine D{\"u}nner, Manolis Sifalakis,
  Haralampos Pozidis, Vasileios Vasileiadis, Michail Vlachos, Cesar Berrospi,
  and Abdel Labbi. 2017.
\newblock Linear-complexity relaxed word mover's distance with gpu
  acceleration.
\newblock In \emph{2017 IEEE International Conference on Big Data (Big Data)},
  pages 889--896. IEEE.

\bibitem[{Balikas et~al.(2018)Balikas, Laclau, Redko, and
  Amini}]{balikas2018cross}
Georgios Balikas, Charlotte Laclau, Ievgen Redko, and Massih-Reza Amini. 2018.
\newblock Cross-lingual document retrieval using regularized wasserstein
  distance.
\newblock In \emph{European Conference on Information Retrieval}, pages
  398--410. Springer.

\bibitem[{Buck and Koehn(2016{\natexlab{a}})}]{buck2016findings}
Christian Buck and Philipp Koehn. 2016{\natexlab{a}}.
\newblock Findings of the wmt 2016 bilingual document alignment shared task.
\newblock In \emph{Proceedings of the First Conference on Machine Translation:
  Volume 2, Shared Task Papers}, pages 554--563.

\bibitem[{Buck and Koehn(2016{\natexlab{b}})}]{buck2016quick}
Christian Buck and Philipp Koehn. 2016{\natexlab{b}}.
\newblock Quick and reliable document alignment via tf/idf-weighted cosine
  distance.
\newblock In \emph{Proceedings of the First Conference on Machine Translation:
  Volume 2, Shared Task Papers}, pages 672--678.

\bibitem[{Chen and Nie(2000)}]{chen2000parallel}
Jiang Chen and Jian-Yun Nie. 2000.
\newblock Parallel web text mining for cross-language ir.
\newblock In \emph{Content-Based Multimedia Information Access-Volume 1}, pages
  62--77. LE CENTRE DE HAUTES ETUDES INTERNATIONALES D'INFORMATIQUE
  DOCUMENTAIRE.

\bibitem[{Clark et~al.(2019)Clark, Celikyilmaz, and Smith}]{clark2019sentence}
Elizabeth Clark, Asli Celikyilmaz, and Noah~A Smith. 2019.
\newblock Sentence mover's similarity: Automatic evaluation for multi-sentence
  texts.
\newblock In \emph{Proceedings of the 57th Annual Meeting of the Association
  for Computational Linguistics}, pages 2748--2760.

\bibitem[{Dara and Lin(2016)}]{dara2016yoda}
Aswarth~Abhilash Dara and Yiu-Chang Lin. 2016.
\newblock Yoda system for wmt16 shared task: Bilingual document alignment.
\newblock In \emph{Proceedings of the First Conference on Machine Translation:
  Volume 2, Shared Task Papers}, pages 679--684.

\bibitem[{Do et~al.(2009)Do, Le, Bigi, Besacier, and Castelli}]{do2009mining}
Thi-Ngoc-Diep Do, Viet-Bac Le, Brigitte Bigi, Laurent Besacier, and Eric
  Castelli. 2009.
\newblock Mining a comparable text corpus for a vietnamese-french statistical
  machine translation system.
\newblock In \emph{Proceedings of the Fourth Workshop on Statistical Machine
  Translation}, pages 165--172. Association for Computational Linguistics.

\bibitem[{El-Kishky et~al.(2019)El-Kishky, Chaudhary, Guzman, and
  Koehn}]{el2019massive}
Ahmed El-Kishky, Vishrav Chaudhary, Francisco Guzman, and Philipp Koehn. 2019.
\newblock A massive collection of cross-lingual web-document pairs.
\newblock \emph{arXiv preprint arXiv:1911.06154}.

\bibitem[{Espl{\`a}-Gomis et~al.(2016)Espl{\`a}-Gomis, Forcada, Rojas, and
  Ferr{\'a}ndez-Tordera}]{espla2016bitextor}
Miquel Espl{\`a}-Gomis, Mikel Forcada, Sergio~Ortiz Rojas, and Jorge
  Ferr{\'a}ndez-Tordera. 2016.
\newblock Bitextor's participation in wmt'16: shared task on document
  alignment.
\newblock In \emph{Proceedings of the First Conference on Machine Translation:
  Volume 2, Shared Task Papers}, pages 685--691.

\bibitem[{Fung and Yee(1998)}]{fung1998ir}
Pascale Fung and Lo~Yuen Yee. 1998.
\newblock An ir approach for translating new words from nonparallel, comparable
  texts.
\newblock In \emph{COLING 1998 Volume 1: The 17th International Conference on
  Computational Linguistics}.

\bibitem[{Gomes and Lopes(2016)}]{gomes2016first}
Lu{\'\i}s Gomes and Gabriel~Pereira Lopes. 2016.
\newblock First steps towards coverage-based document alignment.
\newblock In \emph{Proceedings of the First Conference on Machine Translation:
  Volume 2, Shared Task Papers}, pages 697--702.

\bibitem[{Guo et~al.(2019)Guo, Yang, Stevens, Cer, Ge, Sung, Strope, and
  Kurzweil}]{guo-EtAl:2019:WMT2}
Mandy Guo, Yinfei Yang, Keith Stevens, Daniel Cer, Heming Ge, Yun-hsuan Sung,
  Brian Strope, and Ray Kurzweil. 2019.
\newblock \href {http://www.aclweb.org/anthology/W19-5207} {Hierarchical
  document encoder for parallel corpus mining}.
\newblock In \emph{Proceedings of the Fourth Conference on Machine
  Translation}, pages 64--72, Florence, Italy. Association for Computational
  Linguistics.

\bibitem[{Huang et~al.(2016)Huang, Guo, Kusner, Sun, Sha, and
  Weinberger}]{huang2016supervised}
Gao Huang, Chuan Guo, Matt~J Kusner, Yu~Sun, Fei Sha, and Kilian~Q Weinberger.
  2016.
\newblock Supervised word mover's distance.
\newblock In \emph{Advances in Neural Information Processing Systems}, pages
  4862--4870.

\bibitem[{Jakubina and Langlais(2016)}]{jakubina2016bad}
Laurent Jakubina and Phillippe Langlais. 2016.
\newblock Bad luc@ wmt 2016: a bilingual document alignment platform based on
  lucene.
\newblock In \emph{Proceedings of the First Conference on Machine Translation:
  Volume 2, Shared Task Papers}, pages 703--709.

\bibitem[{Kendall(1938)}]{kendall1938new}
Maurice~G Kendall. 1938.
\newblock A new measure of rank correlation.
\newblock \emph{Biometrika}, 30(1/2):81--93.

\bibitem[{Koehn(2005)}]{koehn2005europarl}
Philipp Koehn. 2005.
\newblock Europarl: A parallel corpus for statistical machine translation.
\newblock In \emph{MT summit}, volume~5, pages 79--86.

\bibitem[{Koehn et~al.(2002)}]{koehn2002europarl}
Philipp Koehn et~al. 2002.
\newblock Europarl: A multilingual corpus for evaluation of machine
  translation.

\bibitem[{Kusner et~al.(2015)Kusner, Sun, Kolkin, and
  Weinberger}]{kusner2015word}
Matt Kusner, Yu~Sun, Nicholas Kolkin, and Kilian Weinberger. 2015.
\newblock From word embeddings to document distances.
\newblock In \emph{International conference on machine learning}, pages
  957--966.

\bibitem[{Lample and Conneau(2019)}]{lample2019cross}
Guillaume Lample and Alexis Conneau. 2019.
\newblock Cross-lingual language model pretraining.
\newblock \emph{arXiv preprint arXiv:1901.07291}.

\bibitem[{Medve{\v{d}} et~al.(2016)Medve{\v{d}}, Jakub{\'\i}cek, and
  Kov{\'a}r}]{medvevd2016english}
Marek Medve{\v{d}}, Milo{\v{s}} Jakub{\'\i}cek, and Vojtech Kov{\'a}r. 2016.
\newblock English-french document alignment based on keywords and statistical
  translation.
\newblock In \emph{Proceedings of the First Conference on Machine Translation:
  Volume 2, Shared Task Papers}, pages 728--732.

\bibitem[{Munkres(1957)}]{munkres1957algorithms}
James Munkres. 1957.
\newblock Algorithms for the assignment and transportation problems.
\newblock \emph{Journal of the society for industrial and applied mathematics},
  5(1):32--38.

\bibitem[{Munteanu and Marcu(2005)}]{munteanu2005improving}
Dragos~Stefan Munteanu and Daniel Marcu. 2005.
\newblock Improving machine translation performance by exploiting non-parallel
  corpora.
\newblock \emph{Computational Linguistics}, 31(4):477--504.

\bibitem[{Munteanu and Marcu(2006)}]{munteanu2006extracting}
Dragos~Stefan Munteanu and Daniel Marcu. 2006.
\newblock Extracting parallel sub-sentential fragments from non-parallel
  corpora.
\newblock In \emph{Proceedings of the 21st International Conference on
  Computational Linguistics and the 44th annual meeting of the Association for
  Computational Linguistics}, pages 81--88. Association for Computational
  Linguistics.

\bibitem[{Papavassiliou et~al.(2016)Papavassiliou, Prokopidis, and
  Piperidis}]{papavassiliou2016ilsp}
Vassilis Papavassiliou, Prokopis Prokopidis, and Stelios Piperidis. 2016.
\newblock The ilsp/arc submission to the wmt 2016 bilingual document alignment
  shared task.
\newblock In \emph{Proceedings of the First Conference on Machine Translation:
  Volume 2, Shared Task Papers}, pages 733--739.

\bibitem[{Rafalovitch et~al.(2009)Rafalovitch, Dale
  et~al.}]{rafalovitch2009united}
Alexandre Rafalovitch, Robert Dale, et~al. 2009.
\newblock United nations general assembly resolutions: A six-language parallel
  corpus.
\newblock In \emph{Proceedings of Machine Translation Summit XII}.

\bibitem[{Ramos et~al.(2003)}]{ramos2003using}
Juan Ramos et~al. 2003.
\newblock Using tf-idf to determine word relevance in document queries.
\newblock In \emph{Proceedings of the first instructional conference on machine
  learning}, volume 242, pages 133--142. Piscataway, NJ.

\bibitem[{Rapp(1999)}]{rapp1999automatic}
Reinhard Rapp. 1999.
\newblock Automatic identification of word translations from unrelated english
  and german corpora.
\newblock In \emph{Proceedings of the 37th annual meeting of the Association
  for Computational Linguistics on Computational Linguistics}, pages 519--526.
  Association for Computational Linguistics.

\bibitem[{Resnik(1999)}]{resnik1999mining}
Philip Resnik. 1999.
\newblock Mining the web for bilingual text.
\newblock In \emph{Proceedings of the 37th annual meeting of the Association
  for Computational Linguistics on Computational Linguistics}, pages 527--534.
  Association for Computational Linguistics.

\bibitem[{Resnik and Smith(2003)}]{resnik2003web}
Philip Resnik and Noah~A Smith. 2003.
\newblock The web as a parallel corpus.
\newblock \emph{Computational Linguistics}, 29(3):349--380.

\bibitem[{Robertson(2004)}]{robertson2004understanding}
Stephen Robertson. 2004.
\newblock Understanding inverse document frequency: on theoretical arguments
  for idf.
\newblock \emph{Journal of documentation}, 60(5):503--520.

\bibitem[{Rubner et~al.(1998)Rubner, Tomasi, and Guibas}]{rubner1998metric}
Yossi Rubner, Carlo Tomasi, and Leonidas~J Guibas. 1998.
\newblock A metric for distributions with applications to image databases.
\newblock In \emph{Sixth International Conference on Computer Vision (IEEE Cat.
  No. 98CH36271)}, pages 59--66. IEEE.

\bibitem[{Shchukin et~al.(2016)Shchukin, Khristich, and
  Galinskaya}]{shchukin2016word}
Vadim Shchukin, Dmitry Khristich, and Irina Galinskaya. 2016.
\newblock Word clustering approach to bilingual document alignment (wmt 2016
  shared task).
\newblock In \emph{Proceedings of the First Conference on Machine Translation:
  Volume 2, Shared Task Papers}, pages 740--744.

\bibitem[{Udupa et~al.(2009)Udupa, Saravanan, Kumaran, and
  Jagarlamudi}]{udupa2009mint}
Raghavendra Udupa, K~Saravanan, A~Kumaran, and Jagadeesh Jagarlamudi. 2009.
\newblock Mint: A method for effective and scalable mining of named entity
  transliterations from large comparable corpora.
\newblock In \emph{Proceedings of the 12th Conference of the European Chapter
  of the Association for Computational Linguistics}, pages 799--807.
  Association for Computational Linguistics.

\bibitem[{Ziemski et~al.(2016)Ziemski, Junczys-Dowmunt, and
  Pouliquen}]{ziemski2016united}
Micha{\l} Ziemski, Marcin Junczys-Dowmunt, and Bruno Pouliquen. 2016.
\newblock The {U}nited {N}ations parallel corpus v1. 0.
\newblock In \emph{Proceedings of the Tenth International Conference on
  Language Resources and Evaluation (LREC 2016)}, pages 3530--3534.

\end{thebibliography}
\bibliographystyle{acl_natbib}

\end{document}